\documentclass[10pt,twocolumn,letterpaper]{article}

\usepackage{iccv}
\usepackage{times}
\usepackage{epsfig}
\usepackage{graphicx}
\usepackage{amsmath}
\usepackage{amssymb}
\usepackage{amsthm}
\usepackage{subfigure}
\usepackage{caption}
\usepackage{enumerate}
\usepackage{algorithm}
\usepackage{multirow}
\usepackage{algorithmic}
\usepackage{color, colortbl}

\algsetup{linenosize=\small}
\pdfoutput=1

\usepackage[pagebackref=false,breaklinks=true,letterpaper=true,colorlinks,bookmarks=false]{hyperref}

\iccvfinalcopy 

\def\iccvPaperID{3144} 

\definecolor{Gray}{gray}{0.9}
\begin{document}

\title{Dense Depth Estimation of a Complex Dynamic Scene without Explicit 3D Motion Estimation}

\author{Suryansh Kumar$^{1}$\quad Ram Srivatsav Ghorakavi$^{1}$\quad Yuchao Dai$^{2}$\quad Hongdong Li$^{1, 3}$\\
Australian National University${^1}$, Northwestern Polytechnical University, China$^2$ \\ Australian Centre for Robotic Vision$^3$
\\
{\tt\small suryansh.kumar@anu.edu.au}\\
}
\maketitle

\begin{abstract}
Recent geometric methods need reliable estimates of 3D motion parameters to procure accurate dense depth map of a complex dynamic scene from monocular images \cite{kumar2017monocular, ranftl2016dense}. Generally, to estimate \textbf{precise} measurements of relative 3D motion parameters and to validate its accuracy using image data is a challenging task. In this work, we propose an alternative approach that circumvents the 3D motion estimation requirement to obtain a dense depth map of a dynamic scene. Given per-pixel optical flow correspondences between two consecutive frames and, the sparse depth prior for the reference frame, we show that, we can effectively recover the dense depth map for the successive frames without solving for 3D motion parameters.  Our method assumes a piece-wise planar model of a dynamic scene, which undergoes rigid transformation locally, and as-rigid-as-possible transformation globally between two successive frames. Under our assumption, we can avoid the explicit estimation of 3D rotation and translation to estimate scene depth. In essence, our formulation provides an unconventional way to think and recover the dense depth map of a complex dynamic scene which is incremental and motion free in nature. Our proposed method does not make object level or any other high-level prior assumption about the dynamic scene, as a result, it is applicable to a wide range of scenarios.  Experimental results on the benchmarks dataset show the competence of our approach for multiple frames.
\end{abstract}

\section{Introduction}\label{ss:intro}
Dense depth estimation of a complex dynamic scene from monocular images is an important and well-studied problem in computer vision \cite{szeliski2010computer}. Recent developments in this area have gained great attention from several industries involved in augmented reality, autonomous driving, movies, robotics \cite{kumar2014markov, kumar2012bayes} \etc. Despite the recent research in solving this problem has provided some promising theory and results, its success depends on the \emph{accurate} estimates of 3D motion parameters \cite{kumar2017monocular, ranftl2016dense, zhou2017unsupervised, godard2018digging}. 

\begin{figure}[t]
\begin{center}
\includegraphics[width=1.0\linewidth]{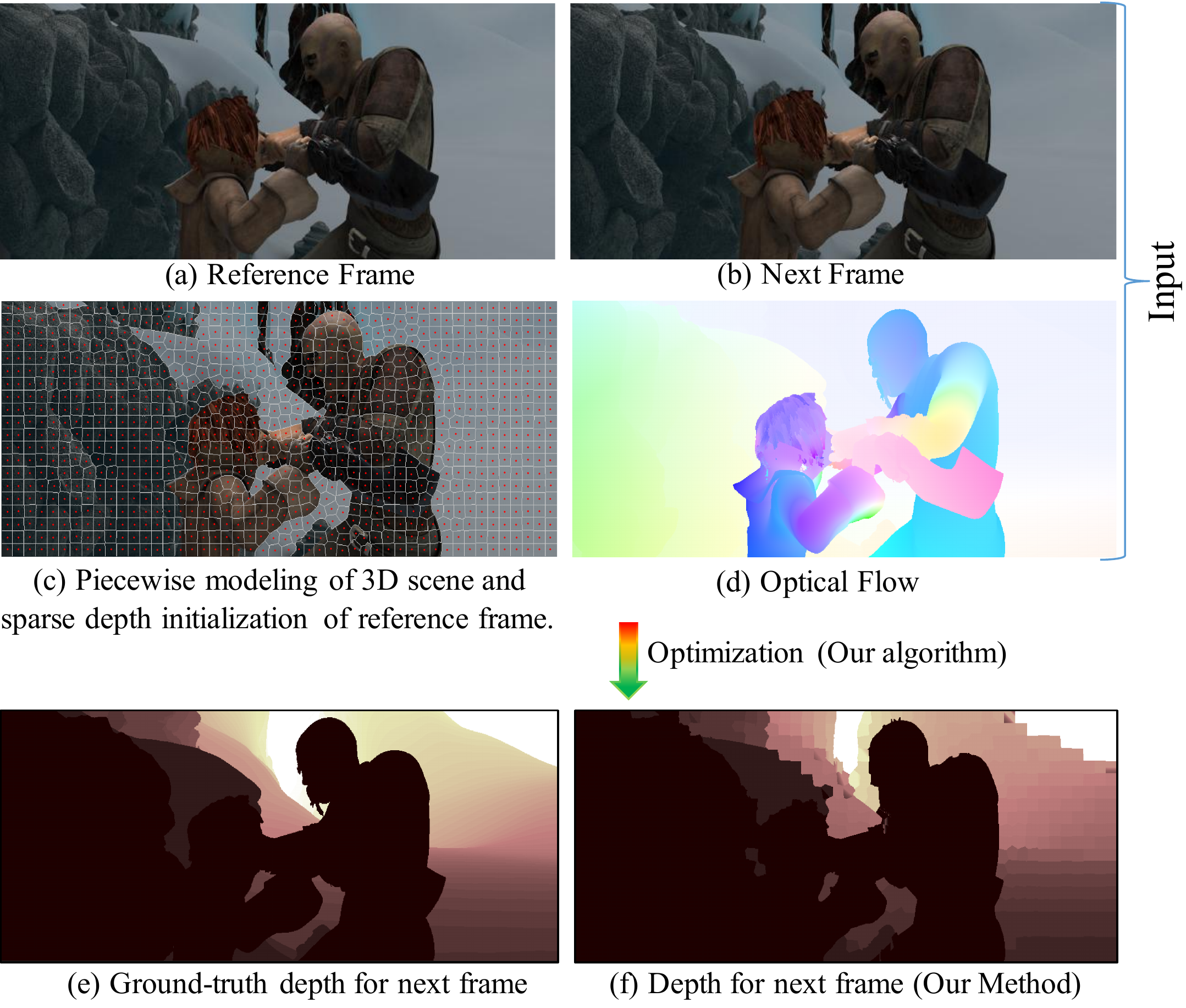}
\caption{\small{Given consecutive monocular perspective frame (a), (b) of a complex dynamic scene and the dense optical flow correspondences between them (d). Also, assume an approximate sparse depth prior for the reference frame is provided as input (c), then, our algorithm under the piecewise planar approximation of a dynamic scene gives per-pixel depth estimate for the next frame (f) without solving for any motion parameters. (e) ground-truth depth.} \label{fig:firstPageresult}}
\end{center}
\end{figure}
To our knowledge, almost all the existing \emph{geometric} solutions to this problem have tried to fit the well-established theory of rigid reconstruction in some way to solve per-pixel depth of \emph{dynamic} scenes \cite{noraky2018depth, kumar2017monocular, ranftl2016dense}. Hence, these extensions are intricate to execute and largely depends on per-object or per-superpixel \emph{reliable} motion estimates \cite{noraky2018depth, kumar2017monocular, ranftl2016dense, achanta2012slic}. The main issue with the available geometric frameworks is that, even if the depth for the first/reference frame is known, we must solve for per-superpixel or per-object 3D motion to obtain the depth for the next frame. Consequently, the composition of their objective function fails to utilize the depth knowledge and therefore, does not cascade the prior knowledge well. In this work, we argue that in a dynamic scene, if the depth for the reference frame is known then it's not obligatory to estimate 3D motion parameters to obtain the depth for the next frame. Hence, the rationale behind relative motion estimation as an essential paradigm for obtaining the depth of a complex dynamic scene seems rather {optional} under the prior knowledge about the depth of the reference frame and dense optical flow between frames. To endorse our claim, we propose an \textbf{alternative approach} which is easy to implement and allow us to get rid of the intricacy related to the optimization on $\mathbb{SE}$(3) manifold.

We posit that the recent geometric methods to solve this task have been bounded by their inherent dependence on the 3D motion parameters. Consequently, we present a different method to solve dense depth estimation problem of a dynamic scene. Inspired by the recent work \cite{kumar2017monocular}, we model the dynamic scene as a set of locally planar surfaces and constrain the change in the scene to be as-rigid-as-possible (ARAP). Recent work by Kumar \etal \cite{kumar2017monocular} uses local rigidity graph structure to constrain the movement of each local planar structure based on the homography \cite{malis2007deeper} and its inter-frame relative 3D motion. In contrast, we propose that the global ARAP assumption of a dynamic scene may not need explicit 3D motion parameters, and its definition just based on the 3D Euclidean distance metric is a sufficient regularization to supply the depth for the next frame. 
To this point, one may ask ``\emph{Why ARAP assumption for a dynamic scene?}''

Consider a general real-world dynamic scene, the change we observe in the scene between consecutive time frame is not arbitrary, rather it is regular. Hence, if we observe a local transformation closely, it changes rigidly, but the overall transformation that the scene undergoes is non-rigid.  Therefore, to assume that the dynamic scene deforms as rigid as possible seems quite convincing and practically works well for most real-world dynamic scenes.

To realize our intuition, we first decompose the dynamic scene as a collection of moving planes. We consider K-nearest neighbors per superpixel \cite{achanta2012slic} (which is an approximation of a surfel in the projective space) to define our ARAP model. For each superpixel, we choose three points \ie, an anchor point (center of the plane), and two other non-collinear points. Since the depth for the reference frame is assumed to be known (for at least 3 non-collinear points per superpixel), we can estimate per plane normal for the reference frame, but to estimate per plane normal for the next frame, we need depth for at least 3 non-collinear points per plane $\S \ref{ss:pwsm}$. If per-pixel depth for the reference frame is known, then ARAP model can be extended to pixel level without any loss of generality. The only reason for such discrete planar approximation is the computational complexity. 

Our ARAP model defined over planes does not take into account the depth continuity along the boundaries of the planes. We address it in the subsequent step by solving a depth continuity constraint optimization problem using the TRW-S algorithm  \cite{kolmogorov2006convergent} (see Fig.~\ref{ss:intro} for a sample result). In this work, we make the following contributions:
\noindent
\begin{itemize}
\setlength{\itemsep}{0pt}
\setlength{\parskip}{0pt}
\setlength{\parsep}{0pt}
\item We propose an approach to estimate the dense depth map of a complex dynamic scene that circumvents explicit parameterization of the inter-frame 3D motion. We specify as rigid as possible constraint for the depth estimation by expressing length consistency constrain directly on locally neighboring 3D points.
\item  Our algorithm under piece-wise planar and as rigid as possible assumption appropriately encapsulates the behavior of a dynamic scene to estimate per pixel depth.
\item  Although the formulation is shown to work ideally for the classical case of two consecutive frames, its incremental in nature and therefore, it is easy to extend to handle multiple frames without estimating 3D motion parameters. Experimental results over multiple frames show the validity of our claim $\S \ref{ss:experiment}$.
\end{itemize}

\section{Related Work and Our Motivation}
Based on our findings,  Li. H \cite{li2010multi} introduced the first method to directly estimate the 3D structure of a scene without explicitly estimating motion. However, this approach  solves 3D structure of a \textbf{rigid} scene and the formulation can handle few \textbf{sparse} points. Very recently, Ji \etal \cite{ji2017maximizing} extended the Li. H \cite{li2010multi} ``motion-free'' framework to solve \textbf{sparse} 3D structure of a \textbf{single} non-rigidly moving object using multiple frames (M view, N point) \cite{li2010multi}. In contrast, we propose a 3D motion free formulation that provides a \textbf{dense} depth map of the \textbf{entire dynamic scene} over frames by relying on global as-rigid-as-possible assumption. Recently, numerous papers have been published for the dense depth estimation of a dynamic scene from images. However, for brevity, in this paper, we limit our discussion to the recent papers that are motivated \emph{geometrically} to solve this problem, leading to the easy discourse of our contributions.

\begin{figure*}
\centering
\subfigure [\label{fig:concept1}] {\includegraphics[height=0.15\textheight]{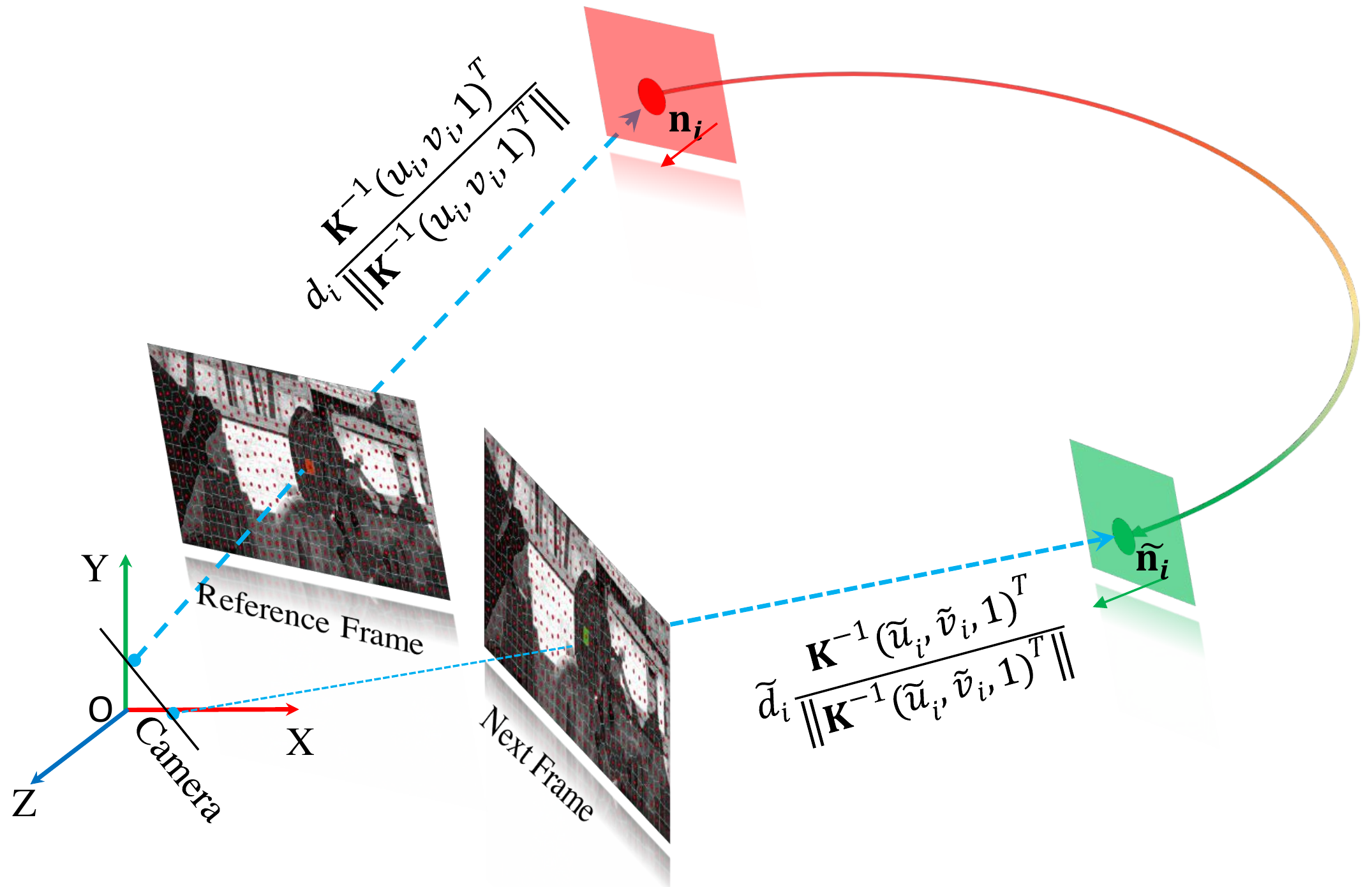}}
~~\subfigure [\label{fig:concept2}] {\includegraphics[height=0.15\textheight]{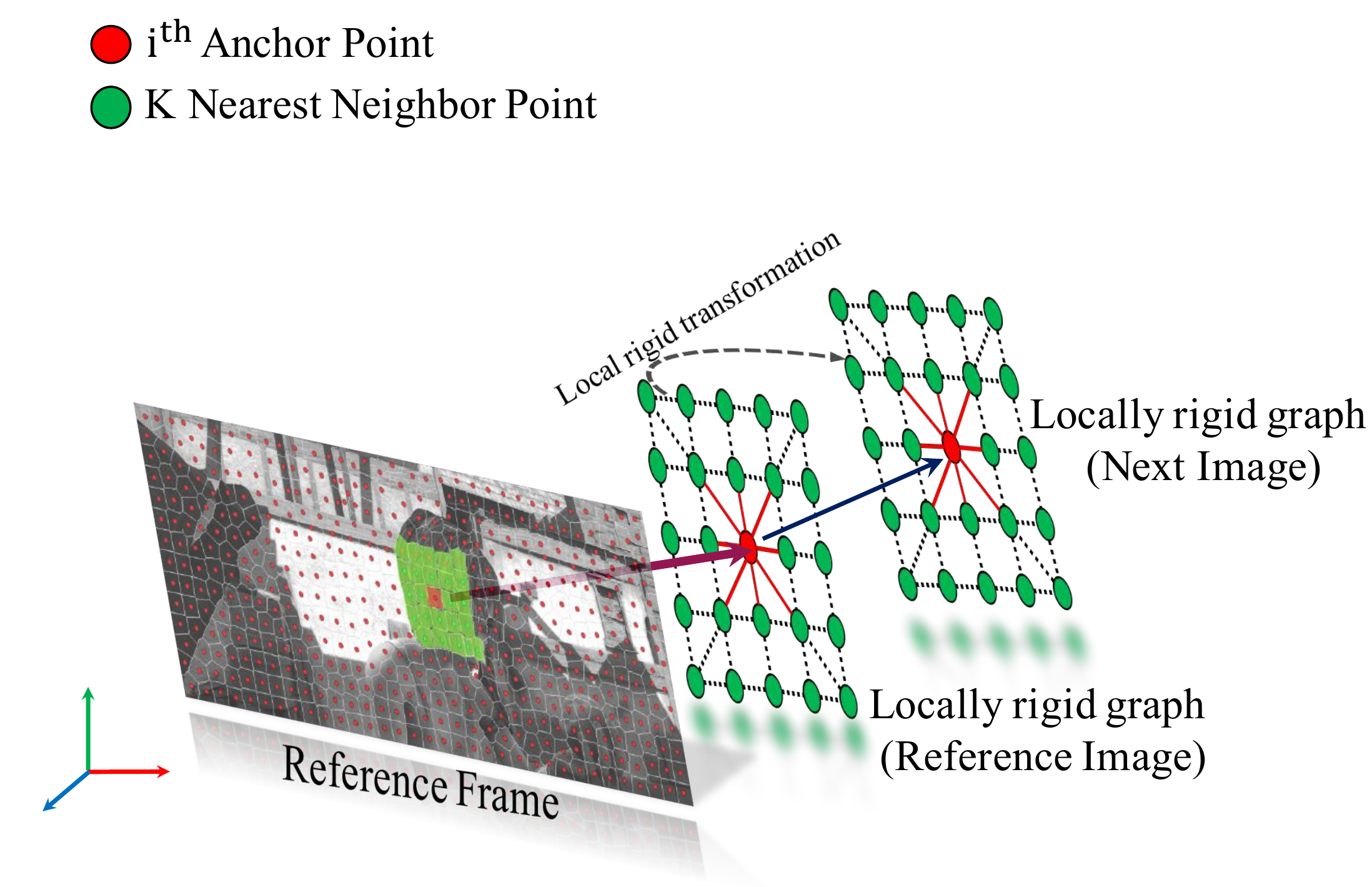}}
\caption{\footnotesize{ (a) Piece-wise planar approximation of a dynamic scene. Each superpixel is assumed to be an approximation of a 3D plane in the projective space. The center of the plane is shown with a filled circle (anchor point). (b) Decomposition of the scene into a local graph structure. Locally rigid graph model with its k-nearest neighbor is shown for the reference frame and the next frame. }}
\label{fig:modelexpl}
\end{figure*}

To the best of our knowledge, two major class of work in the recent past has been proposed for dense depth estimation of an entire dynamic scene from two consecutive monocular images \cite{noraky2018depth, kumar2017monocular, ranftl2016dense}. However, all of these methods depends on explicit 3D motion estimation. These methods can broadly be classified as:

\noindent
{\it{(a) Object-level motion segmentation approach:}} Ranftl \etal \cite{ranftl2016dense} proposed a two/three-staged approach to solve dense monocular depth estimation of a dynamic scene. Given the dense optical flow field, the method first performs an object level motion segmentation using epipolar geometry \cite{hartley2003multiple}. Per-object motion segmentation is then used to perform object-level 3D reconstruction using triangulation \cite{hartley2003multiple}. To obtain a scene consistent depth map, ordering constraint and smoothness constraint were employed over Quick-shift superpixel \cite{vedaldi2008quick} graph to deliver the final result.

\noindent
{\it{(b) Object-level motion segmentation free approach:}} Kumar \etal \cite{kumar2017monocular} argued that ``in a general dynamic scene setting, the task of densely segmenting rigidly moving object or parts is not trivial''. They proposed an over-parametrized algorithm to solve this task without using object-specific motion segmentation. The method dubbed as ``Superpixel Soup'' showed that under two mild assumption about the dynamic scene \ie, (a) the deformation of the scene is locally rigid and globally as rigid as possible and (b) the scene can be approximated by piece-wise planar model, scale consistent 3D reconstruction of a dynamic scene can be obtained for both the frames with a higher accuracy. Inspired by locally rigid assumption, recently, Noraky \etal \cite{noraky2018depth} proposed a method that uses optical flow and depth prior to estimate pose and 3D reconstruction of a deformable object.

\noindent
{\bf{Challenges with such geometric approaches:}} Although these methods provide a plausible direction to solve this problem, its usage to real-world applications can be challenging \cite{ranftl2016dense, kumar2017monocular, noraky2018depth}. The major challenge with these approaches is to \textbf{correctly} estimate all conceivable 3D motion parameters from image correspondences. The method proposed by Ranftl \etal \cite{ranftl2016dense} estimates per-object relative rigid motion which is not a sensible choice if the object themselves are deforming. On the other hand method such as \cite{noraky2018depth, kumar2017monocular} estimates per superpixel/region relative rigid motion which is sensitive to the size of the superpixels and distance of the surfel from the camera.

The point we are trying to make is, given the depth for the reference frame of a dynamic scene, \emph{can we correctly estimate the depth for the next frame using the aforementioned approaches?}. Maybe yes, but then, we have to again estimate relative rigid motion for each object or superpixel and so on and so forth. Inspired by the ``as-rigid-as-possible'' (ARAP) intuition \cite{kumar2017monocular}, in this work, we show that if we know the depth for the reference frame and dense optical flow correspondences between the consecutive frames, then estimating relative 3D motion can be avoided. We can successfully estimate the depth for the next frame by exploiting as-rigid-as-possible global constraint. These depth estimate obtained using ARAP can further be refined using boundary depth continuity constraint. 

The next concern could be \emph{why we are trying to abort the 3D motion data to solve this problem?} Firstly, as alluded to above, such formulation can help avoid involved optimization on $\mathbb{SE}(3)$ manifold. Secondly, it simplifies the underlying objective function which is relatively neat and easy to solve. Thirdly, it provides a distinct view to think about the behaviour of a dynamic scene which generally pivots around the confusion of structure motion and camera motion and its relative inference from image data. Lastly, it provides the flexibility to solve for depth at a pixel level rather than at an object level or superpixel level which is hard to realize using rotation and translation based approaches \cite{noraky2018depth,kumar2017monocular,ranftl2016dense}. Nevertheless, to reduce the overall computational cost, we stick to optimize our objective function at superpixel level.

\section{Piecewise Planar Scene Model}\label{ss:pwsm}
Inspired by the recent work on dense depth estimation of a general dynamic scene \cite{kumar2017monocular}, our model parameterizes the scene as a collection of piece-wise planar surface, where each local plane is assumed to be moving over frames. The global deformation of the entire scene is assumed to be as rigid as possible. Moreover, we assign the center of each plane (anchor point) to act as a representative for the entire points within that plane (see Fig.\ref{fig:modelexpl}). In addition to the anchor point of each plane, we take two more points from the same plane in such a way that these three points are non-collinear (see Fig.\ref{fig:smoothnessexpl}). This strategy is used to define our as rigid as possible constraint between the reference frame and next frame without using any 3D motion parameters. As the depth for the reference frame and the optical flow between the two successive frames is assumed to be known a priori, each local planar region is described using only four parameters ---normal and depth, instead of nine \cite{kumar2017monocular}.

Our model first assigns each pixel of the reference frame to a superpixel using SLIC algorithm \cite{achanta2012slic} and each of these superpixels then acts as a representative for its 3D plane geometry. Since the global change of the dynamic scene is assumed to be ARAP, the transformation that each plane undergoes from the first frame to the next frame should be as minimum as possible. The solution to global ARAP constraint supply depth for three points per plane in the next frame, which is used to estimate the normal and depth of the plane. The estimated depth and normal of each plane are then used to calculate per pixel depth in the next frame.

Although our algorithm is described for the classical two-frame case, it is easy to extend to the multi-frame case. The energy function we define below is solved in two steps: First, we solve for the depth of each superpixel in the next frame using as rigid as possible constraint. Due to the piece-wise planar approximation of the scene, the overall solution to the depth introduces discontinuity along the boundaries. To remove the blocky artifacts ---due to the discretization of the scene, we smooth the obtained depth along the boundaries of all the estimated 3D plane in the second step using TRWS \cite{kolmogorov2006convergent}. If the ARAP cost function is extended to pixel-level then the boundary continuity constraint can be avoided \cite{hornavcek2014highly}. Nevertheless, over-segmentation of the scene provides a good enough approximation of a dynamic scene and is computationally easy to handle.

\subsection{Model overview}
\noindent
{\bf{Notation:}} We refer two consecutive perspective image $\mathbf{I}$, $\mathbf{I}'$ as the reference frame and next frame respectively. Vectors are represented by bold lowercase letters, for \eg `${\bf{x}}$' and the matrices are represented by bold uppercase letters, for \eg `${\bf{X}}$'. The 1-norm,  2-norm of a vector is denoted as $|.|_1$ and $\|.\|_2$ respectively.

\subsection{As-Rigid-As-Possible (ARAP)}\label{ss:arap}
The idea of ARAP constraint is well known in practice and has been widely used for shape modeling and shape manipulation \cite{igarashi2005rigid}. Recently Kumar \etal \cite{kumar2017monocular} exploited this idea to estimate scale consistent dense 3D structure of a dynamic scene. The motivation to use ARAP constraint in our work is inspired by \cite{kumar2017monocular} idea.

Let ($d_i$, $d_j$) and ($\tilde{d_i}$, $\tilde{d_j}$) be the depth of two neighboring 3D points $i, j$ from the reference coordinate in the consecutive frames. Let $(u_i, v_i, 1)^{T}$, $(u_j, v_j, 1)^{T}$ be its image coordinate in the reference frame and $(\tilde{u_i}, \tilde{v_i}, 1)^{T}$, $(\tilde{u_j}, \tilde{v_j}, 1)^{T}$ be its image coordinate in the next frame. If `$\mathbf{K}$' denotes the intrinsic camera calibration matrix then, ${e_i} = \mathbf{K}^{-1}( u_i, v_i ,1)^{{T}}/\|\mathbf{K}^{-1}( u_i, v_i ,1)^{{T}}\|_2$, ${e_j} = \mathbf{K}^{-1}( u_j, v_j ,1)^{{T}}/\|\mathbf{K}^{-1}( u_j, v_j ,1)^{{T}}\|_2$ is the unit vector in the direction of the $i^{th}, j^{th}$ 3D point respectively for the reference frame. Similarly, the corresponding unit vectors in the next frame is denoted with $\tilde{e_i}, \tilde{e_j}$ (see Fig.~\ref{fig:concept1}). Using these notations, we define the ARAP constraint as:
\begin{equation} \label{eq:1}
\Phi^\textrm{arap} = \sum_{i=1}^{3N} \sum_{j \in \mathcal{N}_i^k} w_{ij}^{(1)} \Big| \underbrace{\|d_i{e_i}- d_j{e_j}\|_2}_\text{reference frame} - \underbrace{\|\tilde{d_i}\tilde{e_i} - \tilde{d_j}\tilde{e_j}\|_2}_\text{next frame}\Big|_1
\end{equation}
Here, $N$ is the total number of planes used to approximate the scene and $\mathcal{N}_i^{k}$ is the `$k$' neighboring planes local to $i^{th}$ superpixel (see Fig.~\ref{fig:concept2}). $ w_{ij}^{(1)}$ is the exponential weight fall off based on the image distance of the points.  $ w_{ij}^{(1)}$ parameter slowly breaks the rigidity constraint if the points are far apart in the image space. This constraint encapsulates our idea that the change in the distance of $i^{th}$ point relative to its local neighbors in the next frame should be as minimum as possible. Note that the summation goes over $3N$ rather than $N$ due the reason discussed in Sec. $\S \ref{ss:intro}$

\begin{figure}[t]
\begin{center}
\includegraphics[width=0.58\linewidth, height=0.42\linewidth]{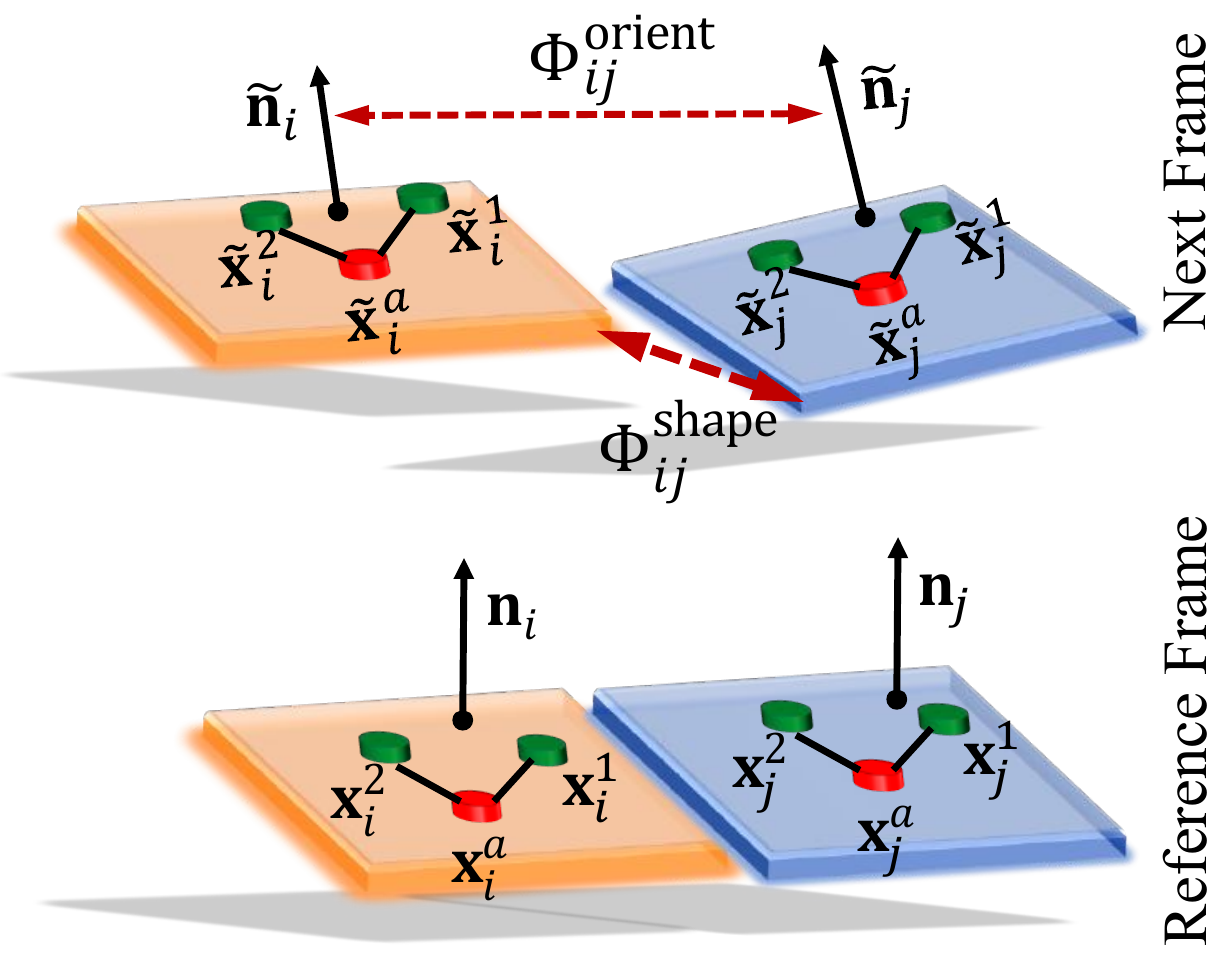}
\caption{\small{Intuition on orientation and shape regularization. Anchor point and two non-collinear points are shown in red and green respectively. Dark red line show the change in the next frame.} \label{fig:smoothnessexpl}}
\end{center}
\end{figure}

\subsection{Orientation and Shape Regularization}\label{ss:osr}
Solving the ARAP constraint provides us the depths for three non-collinear points per-plane for the next frame. We use these three depth estimate per plane to solve for their normals in the next frame. Let the 3D points corresponding to the three depths for $i^{th}$ superpixel in the next frame be denoted as $\tilde{\mathbf{x}}_{i}^{a}$, $\tilde{\mathbf{x}}_{i}^{1}$ and $\tilde{\mathbf{x}}_{i}^{2}$ respectively. We estimate the normals in the next frame as:
\begin{equation}\label{eq:2}
\tilde{\mathbf{n}}_i = \frac{(\tilde{\mathbf{x}}_{i}^{a} - \tilde{\mathbf{x}}_{i}^{1}) \times (\tilde{\mathbf{x}}_{i}^{a} - \tilde{\mathbf{x}}_{i}^{2})}{\|(\tilde{\mathbf{x}}_{i}^{a} - \tilde{\mathbf{x}}_{i}^{1}) \times (\tilde{\mathbf{x}}_{i}^{a} - \tilde{\mathbf{x}}_{i}^{2})\|_2},
\end{equation}
where superscript `$a$' is used intentionally to denote the anchor point, which is assumed to be at the center of each plane (see Fig.~\ref{fig:smoothnessexpl}). Rewriting Eq.~(\ref{eq:2}) in terms of depth
\begin{equation}\label{eq:3}
\tilde{\mathbf{n}}_i = \frac{(\tilde{d}_{i}^{a} \tilde{e}_{i}^a - \tilde{d}_{i}^{1} \tilde{e}_{i}^1) \times (\tilde{d}_{i}^{a} \tilde{e}_{i}^a - \tilde{d}_{i}^{2} \tilde{e}_{i}^2)}{\|(\tilde{d}_{i}^{a} \tilde{e}_{i}^a - \tilde{d}_{i}^{1} \tilde{e}_{i}^1) \times (\tilde{d}_{i}^{a} \tilde{e}_{i}^a - \tilde{d}_{i}^{2} \tilde{e}_{i}^2)\|_2}.
\end{equation}

\noindent
{\bf{(a) Orientation smoothness constraint:}} Once we compute the normal for each plane and 3D coordinates of the anchor point, which lies on the plane, we estimate the depth of the plane as follows
\begin{equation}\label{eq:4}
\tilde{\mathbf{n}}_i^{T}\tilde{\mathbf{x}}_{i}^{a} = \tilde{d}_{i}^{\pi_a}.
\end{equation}
The computed depth of the plane is then used to solve for per-pixel depth in the next frame ---assuming the intrinsic camera matrix is known \cite{kumar2017monocular,hartley2003multiple}.  To encourage the smoothness in the change of angles between each adjacent planes (see Fig.~\ref{fig:smoothnessexpl}), we define the orientation regularization as
\begin{equation}\label{eq:5}
\Phi_{ij}^\textrm{orient} = \lambda_1 \rho_1 \Big(1 - \frac{|\tilde{\mathbf{n}}_{i}^{T}\tilde{\mathbf{n}}_{j}|}{\|\tilde{\mathbf{n}}_{i}\| \|\tilde{\mathbf{n}}_{j}\|}\Big),
\end{equation}
where, $\lambda_1$ is an empirical constant and $\rho_{1}(x)$ = $\min(|x|, \sigma_1)$ is the truncated $l_1$ function with $\sigma_1$ as a scalar parameter.

\noindent
{\bf{(b) Shape smoothness constraint:}} In our representation, the dynamic scene model is approximated by the collection of piecewise planar regions. Hence, the solution to per-pixel depth obtained using Eq.~(\ref{eq:1}) to Eq.~(\ref{eq:4}) may provide discontinuity along the boundaries of the planes in 3D (see Fig.~\ref{fig:smoothnessexpl}). To allow smoothness in the 3D coordinates for each adjacent planes along their region of separation, we define the shape smoothness constraint as
\begin{equation}\label{eq:6}
\begin{aligned}
\Phi^\textrm{shape}=\sum_{(i, j) \in N_{b}} w_{ij}^{(2)}\rho_2(\underbrace{\|d_i{e_i}- d_j{e_j}\|_2^2}_\text{reference frame} + \underbrace{\|\tilde{d_i}\tilde{e_i} - \tilde{d_j}\tilde{e_j}\|_2^2}_\text{next frame}).
\end{aligned}
\end{equation}
\noindent
The symbol `$N_{b}$' denotes the set of boundary pixels of $i^{th}$ superpixel that are shared with the boundary pixel of other superpixels. The weight $w_{ij}^{(2)}$ = $\exp(-\beta \|\mathbf{I}_i - \mathbf{I}_j\|_2)$ takes into account the color consistency of the plane along the boundary points ---weak continuity constraint \cite{blake1987visual}. Since all the pixels within the same plane are assumed to share the same model, smoothness for the pixels within the plane is not essentially required. Similar to orientation regularization, $\rho_{2}(x)$ = $\min(|x|, \sigma_2)$ is the truncated $l_1$ penalty function with $\sigma_2$ as a scalar parameter. The overall optimization steps of our method is provided in {\bf{Algorithm}} ({\bf{\ref{algo:1}}}).

\begin{algorithm}[h!]
\label{Algorithm}
\caption{:~~\small{Dense Depth Estimation without using 3D motion}}
\begin{algorithmic}
{\small
\STATE {\bf{Input:}} ($\mathbf{I}, \mathbf{I}'$), optical\_flow($\mathbf{I}, \mathbf{I}'$), $\mathbf{K}$, depth for reference frame.

\STATE {\bf{Output:}} Dense depth map for the next frame. 

\STATE {\fontfamily{cmtt}\selectfont 1:} Over-segment the reference frame into $N$ superpixels \cite{achanta2012slic}.
\STATE {\fontfamily{cmtt}\selectfont 2:} Assign anchor point for each superpixel and two other points in the same plane such that these three points are non-collinear (see Fig.~\ref{fig:smoothnessexpl}).
 \STATE {\fontfamily{cmtt}\selectfont 3:} Use K-NN algorithm over superpixels to get the K-nearest neighbor index set.
 \STATE {\fontfamily{cmtt}\selectfont 4:}  Solve for per-superpixel depth in the next frame $\S \ref{ss:arap}$
 {\fontfamily{cmtt}\selectfont
 \begin{equation}
 \begin{aligned}
 & \displaystyle  ~~\Phi^{\text{arap}} \rightarrow \underset{\tilde{d}_{i}} {\text{minimize}} \\
 & \displaystyle \text{subject to:} ~~\tilde{d}_{i} > 0,  ~~|\tilde{d}_{i} - d_i|< d_{i\sigma} (\textit{optional}) \\
 & \displaystyle \textrm{where,} ~d_{i\sigma} ~\textrm{is the variance in the depth.}\\
 \end{aligned}\label{eq:7}
 \end{equation}
 }
{\small{{{Note:}} The second constraint provides a trust region for the fast and proper convergence of a non-convex problem (Fig.\ref{fig:statsexperiment2}). Can be thought of as max/min restriction  to the scene deformation.}}
\STATE {\fontfamily{cmtt}\selectfont 5:} Estimate the normal of each plane in the next frame Eq.~(\ref{eq:3}).
\STATE {\fontfamily{cmtt}\selectfont 6:} Estimate the depth of each plane Eq.~(\ref{eq:4}).
\STATE {\fontfamily{cmtt}\selectfont 7:} Solve per pixel depth for the next frame using per plane depth $(\tilde{d}_{i}^{\pi_a})$, $\mathbf{K}$, normal of the plane and its image coordinate.
\STATE {\fontfamily{cmtt}\selectfont 8:} Refine the depth of the next frame by minimizing Eq.~(\ref{eq:5}), Eq.~(\ref{eq:6}) with respect to depth and normal \cite{kolmogorov2006convergent} $\S \ref{ss:osr}$. 
{\fontfamily{cmtt}\selectfont
\begin{equation}
 \begin{aligned}
 & \displaystyle ~~(\Phi^{\text{orient}} + \Phi^{\text{shape}})\rightarrow \underset{\tilde{d}_{i}, \tilde{\mathbf{n}}_i} {\text{minimize}} \\
 & \displaystyle \text{subject to:} ~~\tilde{d}_{i} > 0, \|\tilde{\mathbf{n}}_i\| = 1.
 \end{aligned}\label{eq:8}
 \end{equation}
 }
\STATE {\fontfamily{cmtt}\selectfont 9:} ({\bf{Optional}}) For generalizing the idea to multi-frame, repeat the above steps by making the next frame as the reference frame and new frame as the next frame.}
\end{algorithmic}\label{algo:1}
\end{algorithm}

\section{Experimental Evaluation}\label{ss:experiment}
We performed the experimental evaluation of our approach on two benchmark datasets, namely MPI Sintel \cite{butler2012naturalistic} and KITTI \cite{geiger2013vision}. These two datasets conveniently provide a complex and realistic environment to test and compare our dense depth estimation algorithm. We compared the accuracy of our approach against two recent state-of-the-art methods \cite{kumar2017monocular,ranftl2016dense} that use geometric approach to solve dynamic scene dense depth estimation from monocular images. These comparisons are performed using three different dense optical flow estimation algorithms, namely PWC-Net \cite{Sun2018PWC-Net}, FlowFields \cite{bailer2015flow} and Full Flow \cite{chen2016full}. All the depth estimation accuracies are reported using mean relative error (MRE) metric. Let $\tilde{d}$ be the estimated depth and $\tilde{d}^{gt}$ be the ground-truth depth, then MRE is defined as 
\begin{equation}
\textrm{MRE} = \frac{1}{P} \sum_{i=1}^{P} \frac{|\tilde{d}_{i}-\tilde{d}_{i}^{gt}|}{\tilde{d}_{i}^{gt}},
\end{equation}
where `$P$' denotes the total number of points. The statistical results for DMDE \cite{ranftl2016dense} and Superpixel Soup \cite{kumar2017monocular} are taken from their published work for comparison.

\noindent
\textbf{Implementation Details:} We over-segment the reference frame into 1000-1200 superpixels using SLIC algorithm \cite{achanta2012slic} to approximate the scene. 
We use a fixed value of $d_{i\sigma}$ = 1 and $\mathcal{N}_{i}^{k}$ = 20-25 for all the experiments. For computing the dense optical flow correspondences between images we used  PWC-Net \cite{Sun2018PWC-Net}, FlowFields \cite{bailer2015flow} and Full Flow \cite{chen2016full} algorithm. The depth for the reference image is initialized using Mono-Depth \cite{godard2017unsupervised} model on the KITTI dataset and using Superpixel Soup algorithm \cite{kumar2017monocular} on the MPI-Sintel dataset. The proposed optimization is solved in two stages, firstly Eq.~(\ref{eq:7}) is optimized using SQP \cite{powell1978fast} algorithm and Eq.~(\ref{eq:8}) is optimized using TRW-S \cite{kolmogorov2006convergent} algorithm. The choice of the optimizer is purely empirical, and the user may choose different optimization algorithms to solve the same cost function. We implemented it using C++/MATLAB which takes 10-12 minutes on a commodity desktop computer to provide the result.

The implementation is performed under two different experimental settings. In the first setting, given the sparse (\ie for three non-collinear points per superpixel) depth estimate of a dynamic scene for the reference frame, we estimate the per-pixel depth for the next frame. In the second experimental setting, we generalize this idea of two frame depth estimation to multiple frames by computing the depth estimates over frames. For easy understanding, MATLAB codes are provided in the supplementary material showing our idea of ARAP on synthetic examples of a dynamic scene.

\begin{figure}
\begin{center}
\includegraphics[width=1.0\linewidth, height=0.15\textheight]{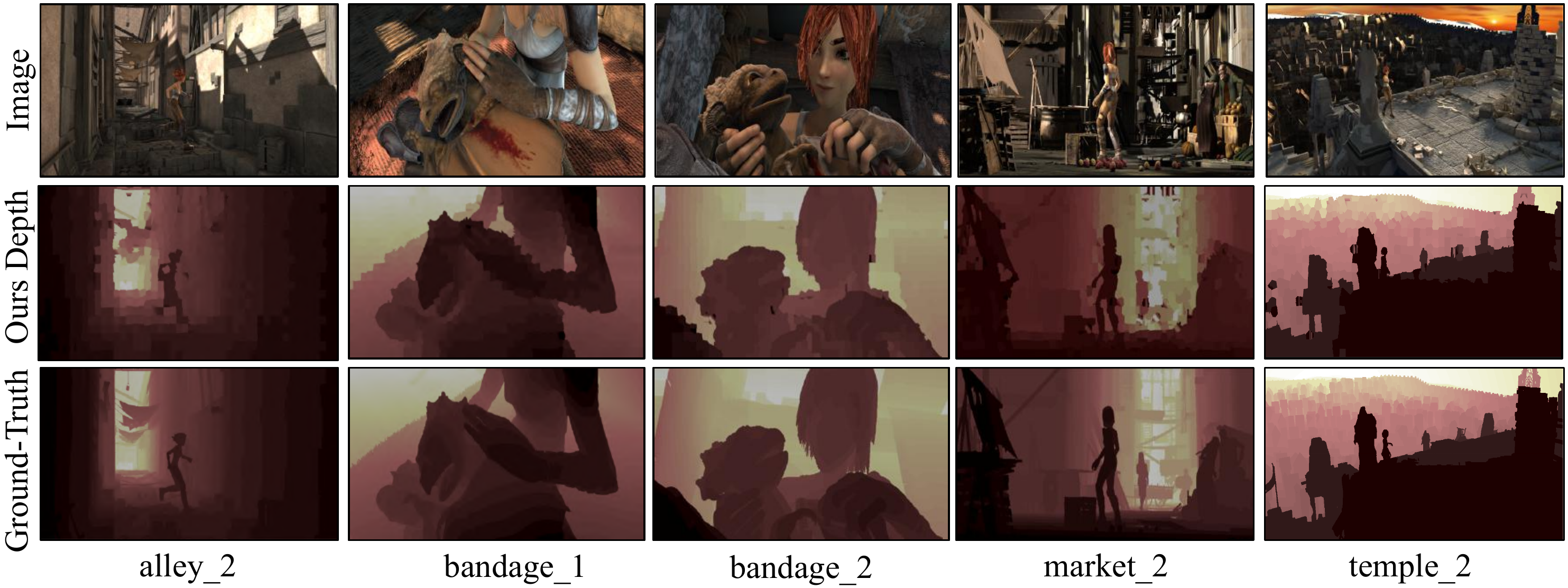}
\caption{\small{Depth results on the MPI Sintel dataset\cite{butler2012naturalistic} for the next frame under two frame experimental setting. $\mathbf{2^{nd}}$ and $\mathbf{3^{rd}}$ row show ours and ground-truth depth map results respectively.} \label{fig:MPItwoframe}}
\end{center}
\end{figure}

\subsection{MPI Sintel}
This dataset gives an ideal setting to evaluate depth estimation algorithms for complex dynamic scenes. It contains image sequences with intricate motions and severe illumination change. Moreover, the large number of non-planar scenes and non-rigid deformations makes it a suitable choice to test the piece-wise planar assumption. We selected seven set of scenes namely alley\_1, alley\_2, ambush\_5, bandage\_1, bandage\_2, market\_2 and temple\_2 from the clean category of this dataset to test our method.

\begin{table}
\centering
\scriptsize
\begin{tabular}{|>{\columncolor[gray]{0.88}}c|c|c|c|c|}
\hline
{\scriptsize{OF}$\downarrow \slash $ \scriptsize{Methods} $\rightarrow$} & {\scriptsize{DMDE \cite{ranftl2016dense}}}  & {\scriptsize{S. Soup \cite{kumar2017monocular}}} & {\scriptsize{Ours}}\\ \hline
PWC Net \cite{Sun2018PWC-Net}  &  -   &  -   & 0.1848  \\ \hline
Flow Fields \cite{bailer2015flow}  &  0.2970   &  0.1669   & 0.1943  \\ \hline
Full Flow  \cite{chen2016full} &   -   &  0.1933   & 0.2144  \\ \hline
\end{tabular}
\caption{ \small{Comparison of dense depth estimation methods under \emph{two consecutive frame setting} against the state-of-the-art approaches on the {\bf{MPI Sintel dataset}} \cite{butler2012naturalistic}. For consistency, the evaluations are performed using mean relative error metric (MRE).}} \label{tab:MPIResults}
\end{table}

\noindent
{\bf{(a) Two-frame results:}}
While testing our algorithm for the two-frame case, the reference frame depth is initialized using recently proposed superpixel-soup algorithm \cite{kumar2017monocular}. The optical flow between the frames is obtained using methods such as PWC-Net \cite{Sun2018PWC-Net}, Flow Fields \cite{bailer2015flow} and Full Flow \cite{chen2016full}.  Table (\ref{tab:MPIResults}) shows the statistical performance comparison of our method against other geometric approaches. The statistics clearly show that our alternative way performs almost equally well without using any 3D rotation or translation.
Qualitative results within this setting are shown in Fig.~\ref{fig:MPItwoframe}.

\begin{figure}
\begin{center}
\includegraphics[width=1.0\linewidth,height=0.15\textheight]{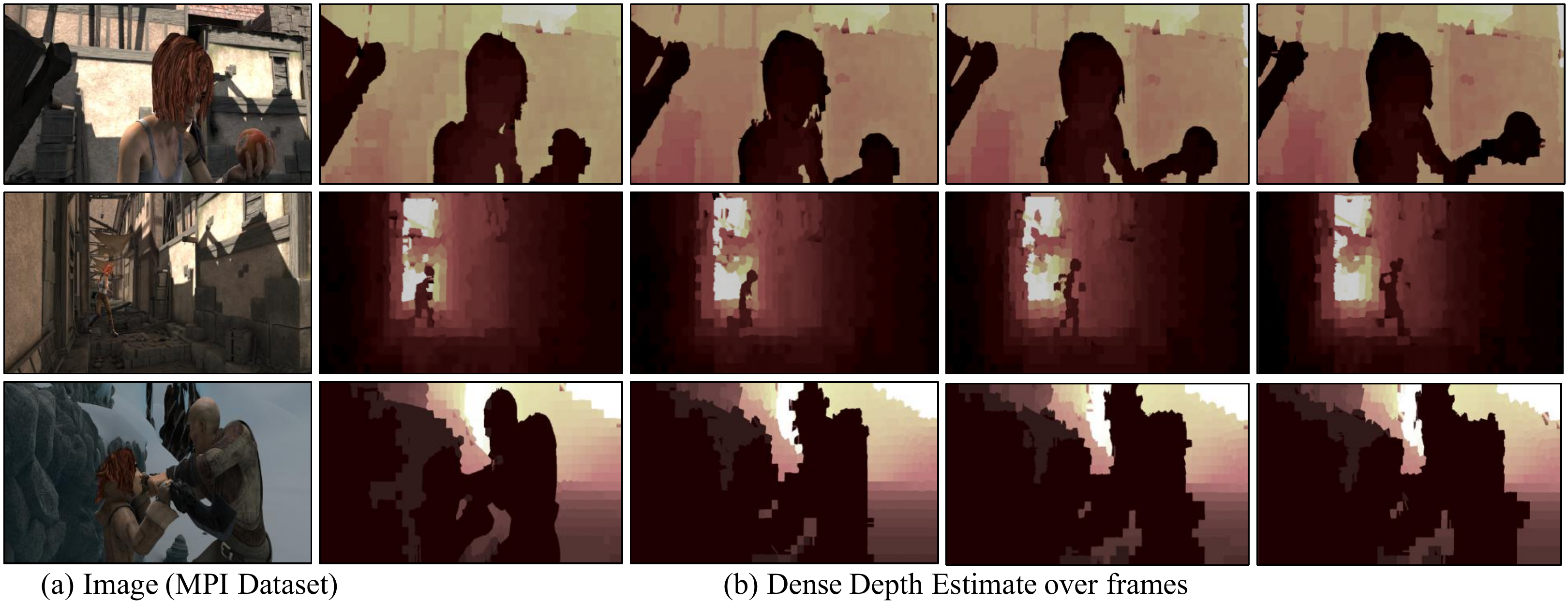}
\caption{\small{Results on MPI Sintel dataset \cite{butler2012naturalistic} under multi-frame experimental setting. (a) Image frame for which the depth is initialized. (b) Depth estimation results using our method over frames.} \label{fig:MPImultiframe}}
\end{center}
\end{figure}

\noindent
{\bf{(b) Multi-frame results:}}
In the multi-frame setting, only the depth for the first frame is initialized. The result obtained for the next frame is then used for the upcoming frames to estimate its dense depth map. Since we are dealing with the dynamic scene, the environment changes slowly and therefore, the error starts to accumulate over frames. Fig.~\ref{fig:n1} reflects this propagation of error over frames. Qualitative results over multiple frames are shown in Fig.~\ref{fig:MPImultiframe}.

\subsection{KITTI}
The KITTI dataset has emerged as a standard benchmark dataset to evaluate the performance of dense depth estimation algorithms. It contains images of outdoor driving scenes with different lighting conditions and large camera motion. We tested our algorithm on both KITTI raw data and KITTI 2015 benchmark. For KITTI dataset, we used Monodepth method \cite{godard2017unsupervised} to initialize the reference frame depth. Dense optical flow correspondences are obtained using the same aforementioned methods. For consistency, the depth estimation error measurement on KITTI dataset follows the same order of 50 meters as presented in \cite{godard2017unsupervised} work.

\begin{figure}
\begin{center}
\includegraphics[width=1.0\linewidth, height=0.15 \textheight]{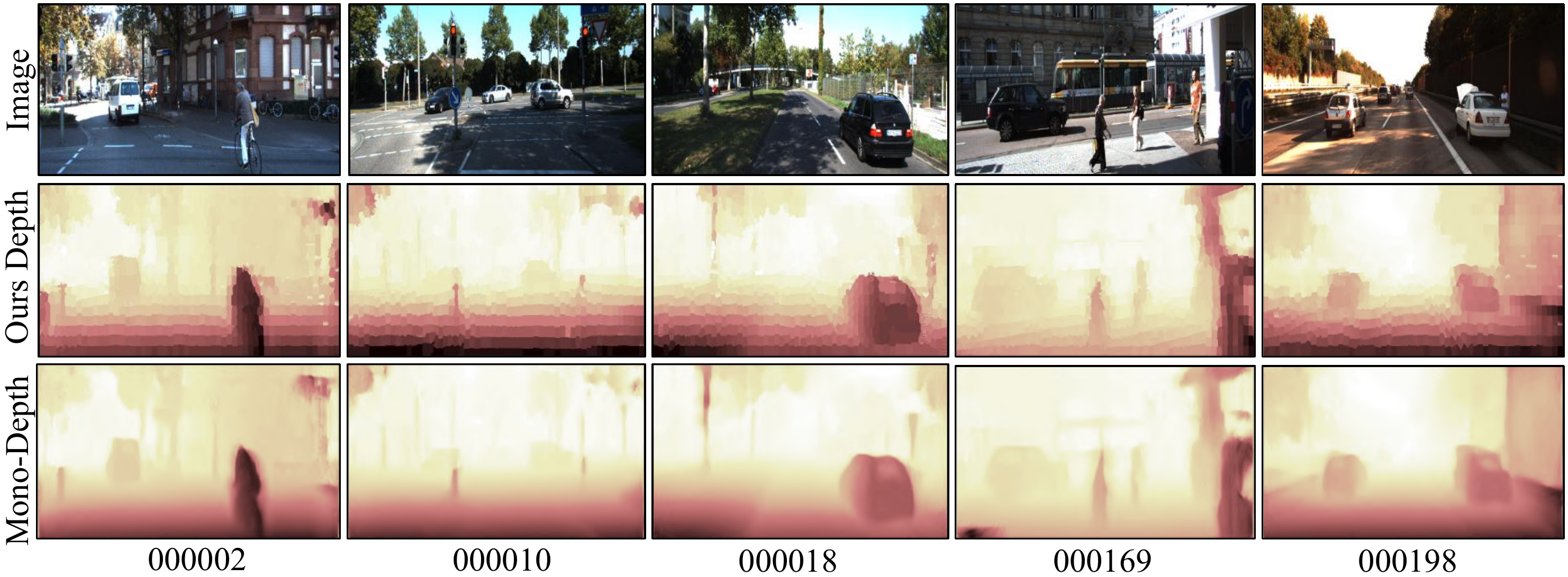}
\caption{\small{Results on KITTI 2015 benchmark dataset under two frame experimental setting. {\bf{$\mathbf{3^{rd}}$ row:}} Monodepth \cite{godard2017unsupervised} results on the same sequence for the next frame for qualitative comparison.} \label{fig:KITTItwoframe}}
\end{center}
\end{figure}

\noindent
{\bf{Two-frame results:}}
KITTI 2015 scene flow dataset provides two consecutive frame pair of a dynamic scene to test algorithms. Table (\ref{tab:KITTIResults}) provides the depth estimation statistical result of our algorithm in comparison to other competing methods. Here, our results are a bit better using PWC-Net \cite{Sun2018PWC-Net} optical flow and Monodepth \cite{godard2017unsupervised} depth initialization.
Fig.~\ref{fig:KITTItwoframe} shows the qualitative results using our approach in comparison to the Monodepth \cite{godard2017unsupervised} for the next frame.

\begin{figure}
\begin{center}
\includegraphics[width=1.0\linewidth, height=0.15\textheight]{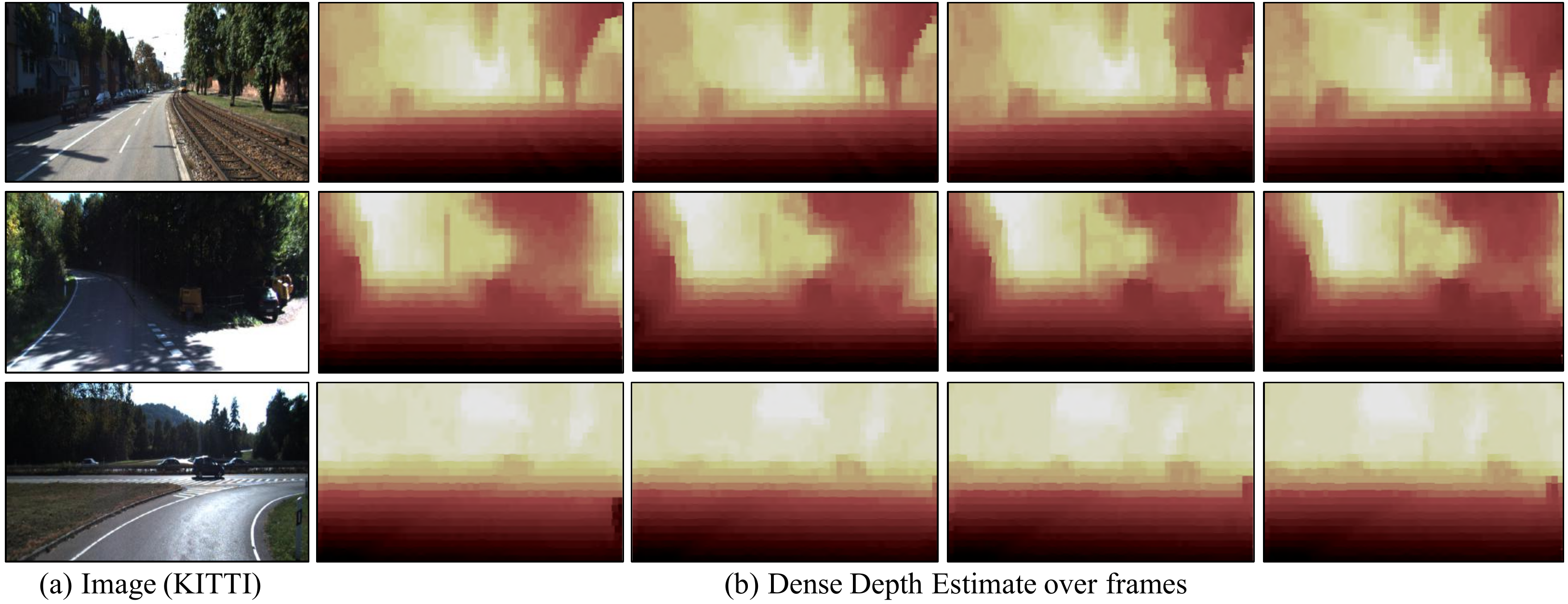}
\caption{\small{Results on KITTI raw dataset under multi-frame experimental setup. (a) Reference image for which the depth is initialized (b) Dense depth results over frames using our algorithm.} \label{fig:KITTImultiframe}}
\end{center}
\end{figure}

\noindent
{\bf{Multi-frame results:}}
To test the performance of our algorithm on multi-frame KITTI dataset, we used KITTI raw dataset specifically from the city, residential and road category. The depth for only the first frame is initialized using Monodepth \cite{godard2017unsupervised} and then we estimate the depth for the upcoming frames. Due to very large displacement in the scene per frame ($>$150) pixels, the rate of change of error accumulation curve for KITTI dataset (Fig.~\ref{fig:n2}) is a bit steeper than MPI Sintel. Fig.~\ref{fig:KITTImultiframe} and Fig.~\ref{fig:n2} show the qualitative results and depth error accumulation over frames on KITTI raw dataset respectively.

\begin{table}
\centering
\scriptsize
\begin{tabular}{|>{\columncolor[gray]{0.88}}c|c|c|c|c|}
\hline
{\scriptsize{OF}$\downarrow \slash $ \scriptsize{Methods} $\rightarrow$} & {\scriptsize{DMDE \cite{ranftl2016dense}}}  & {\scriptsize{S. Soup \cite{kumar2017monocular}}} & {\scriptsize{Ours}}\\ \hline
PWC Net \cite{Sun2018PWC-Net}  &  -   &  -   & 0.1182  \\ \hline
Flow Fields \cite{bailer2015flow}  &  0.1460   &  0.1268   & 0.1372  \\ \hline
Full Flow  \cite{chen2016full} &   -   &  0.1437   & 0.1665  \\ \hline
\end{tabular}
\caption{ \small{Comparison of dense depth estimation under \emph{two consecutive frame setting} against the state-of-the-art approaches on {\bf{KITTI  dataset}} \cite{butler2012naturalistic}. For consistency, the evaluations are performed using mean relative error metric (MRE). The results are better due to monodepth initialization for the reference frame.}} \label{tab:KITTIResults}
\end{table}

\section{Statistical Analysis}
Besides standard experiments under the aforementioned variable initialization, we conducted other experiments to better understand the behavior of the proposed idea. We conducted experiments on a synthetic example shown in Fig.~\ref{fig:syntheticexample} for easy understanding to the readers. Matlab codes are provided in the supplementary material for reference. 

\begin{figure*}
\centering
\subfigure [\label{fig:n1}] {\includegraphics[width=0.23\textwidth, height=0.13\textheight]{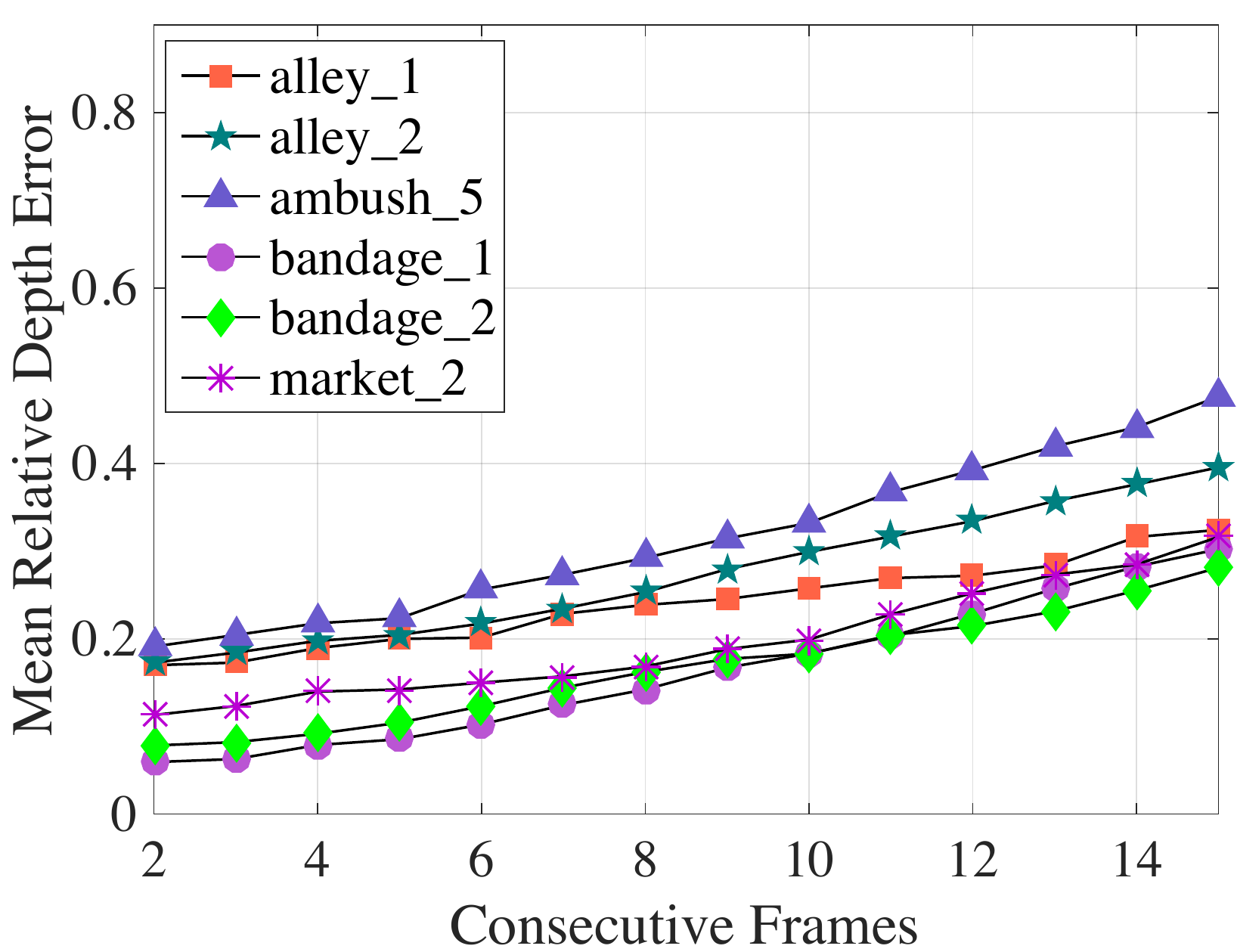}}
\subfigure [\label{fig:n2}] {\includegraphics[width=0.23\textwidth, height=0.13\textheight]{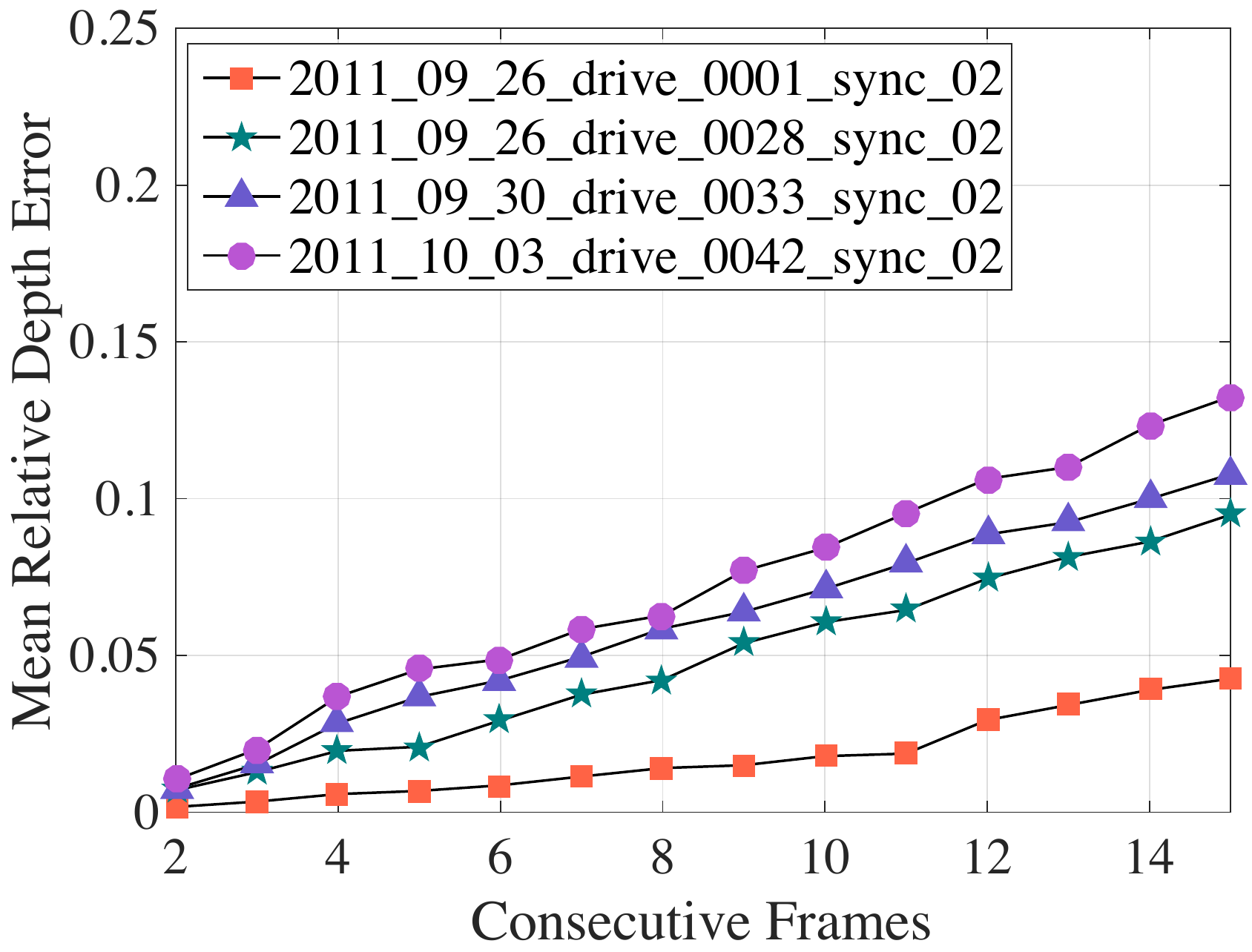}}
\subfigure [\label{fig:m1}] {\includegraphics[width=0.23\textwidth, height=0.13\textheight]{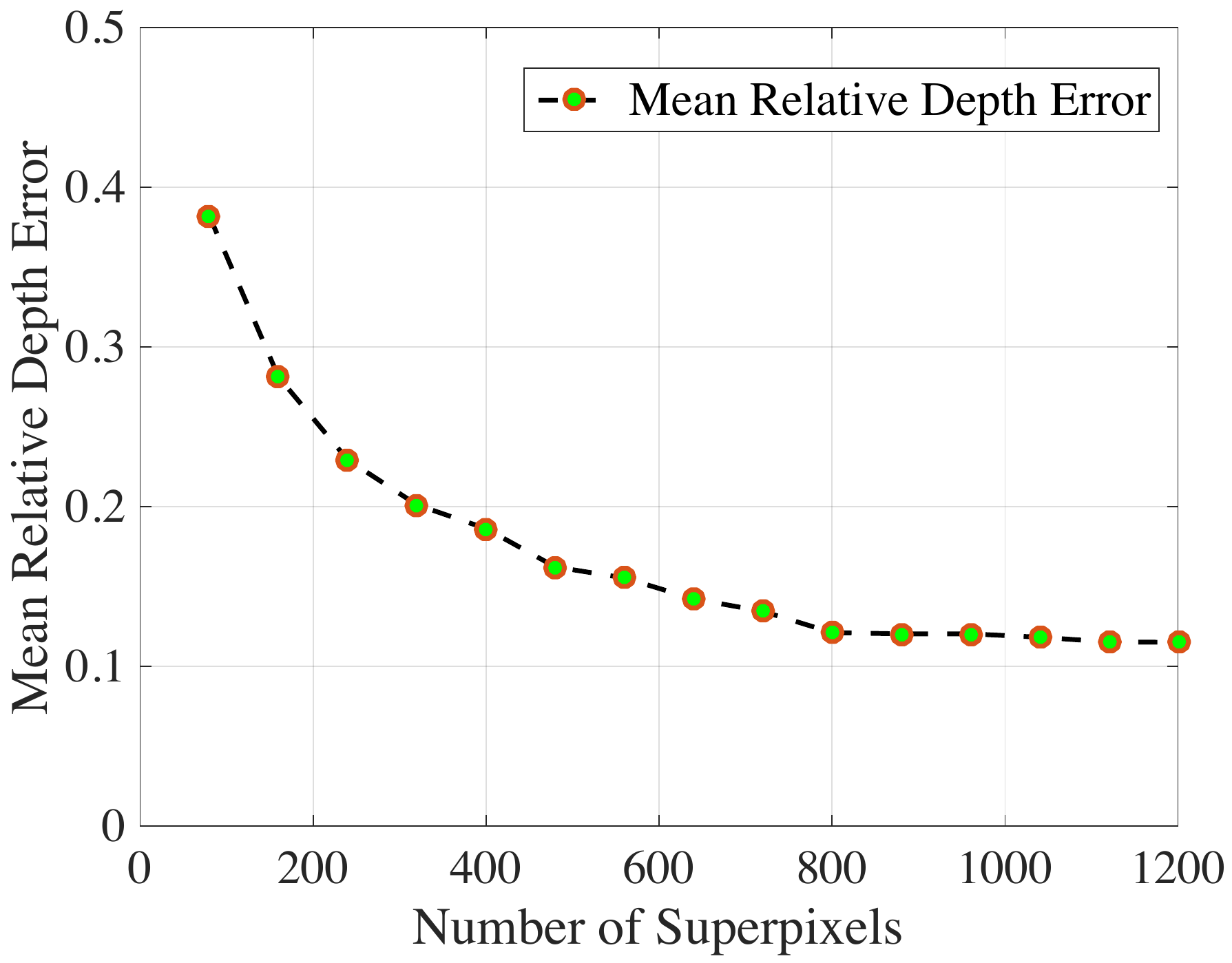}}
\subfigure [\label{fig:o1}] {\includegraphics[width=0.23\textwidth, height=0.13\textheight]{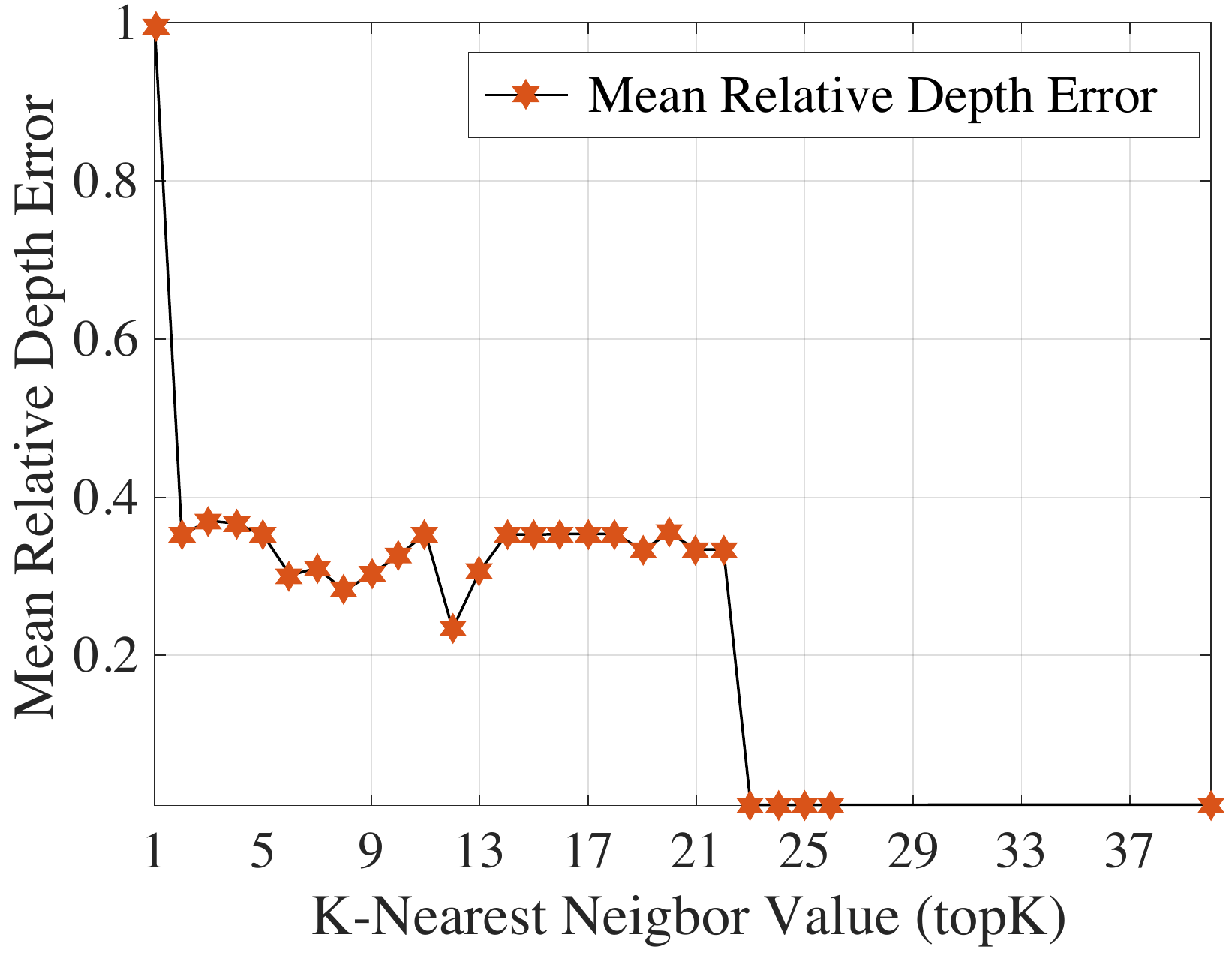}}
\caption{\small{(a)-(b) Accumulation of error over frames for MPI and KITTI dataset respectively. (c) Change in the depth estimation accuracy w.r.t number of superpixel. (d) Variation in the depth accuracy as a function of k-nearest neighbor ($\mathcal{N}_{i}^{k}$)}}
\label{fig:statsexperiment1}
\end{figure*}

\noindent
{\bf{(a) Effect of the variable $N$}}: The number of superpixels to approximate the dynamic scene can affect the performance of our method. A small number of superpixel can provide poor depth result, whereas a very large number of superpixel can increase the computation time. Fig.~\ref{fig:m1} shows the change in the accuracy of depth estimation with respect to the change in the number of superpixels. The curve suggests that for KITTI and MPI Sintel 1000-1200 superpixel provides a reasonable approximation to the dynamic scenes.

\noindent
{\bf{(b) Effect of the variable $\mathcal{N}_{i}^{k}$}}: The number of K-nearest neighbors to define the local rigidity graph can have a direct effect on the performance of the algorithm. Although $\mathcal{N}_{i}^{k} = 20-25$ works well for the tested benchmarks, it is purely an empirical parameter and can be different for  a distinct dynamic scene. Fig.~\ref{fig:o1} demonstrates the performance of the algorithm with the change in the values of $\mathcal{N}_{i}^{k}$.

\begin{figure*}
\centering
\subfigure [\label{fig:sn1}] {\includegraphics[width=0.23\textwidth, height=0.13\textheight]{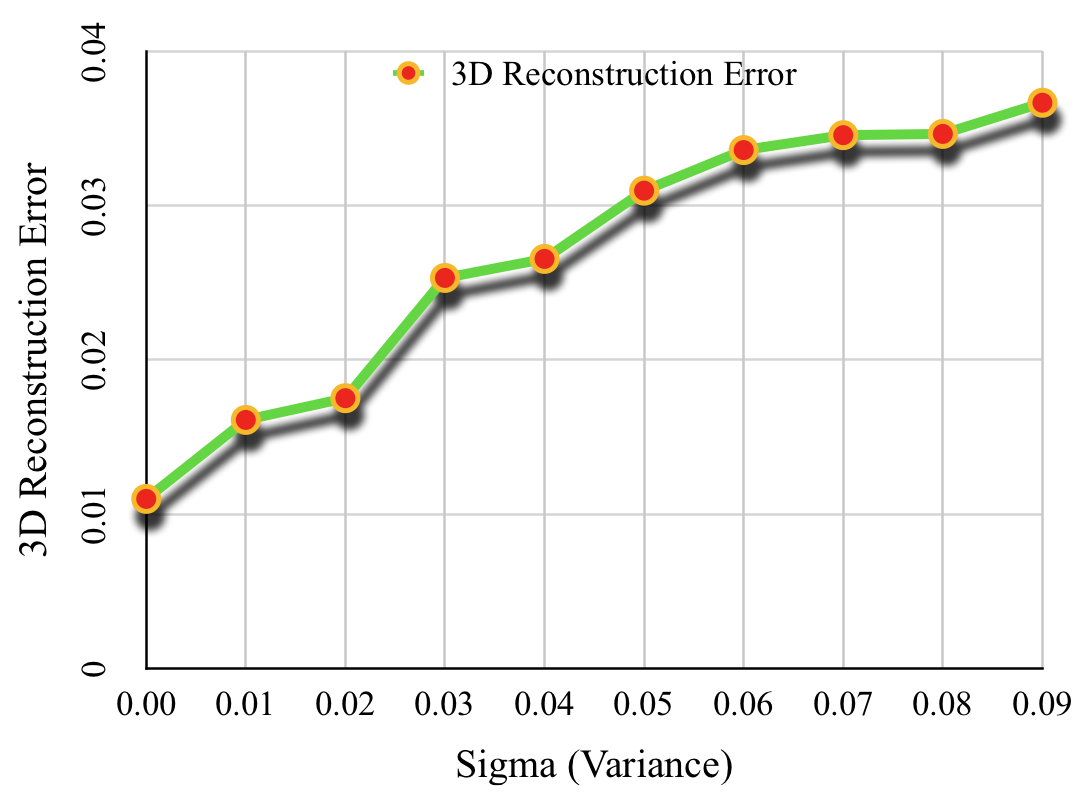}}
\subfigure [\label{fig:sn2}] {\includegraphics[width=0.23\textwidth, height=0.13\textheight]{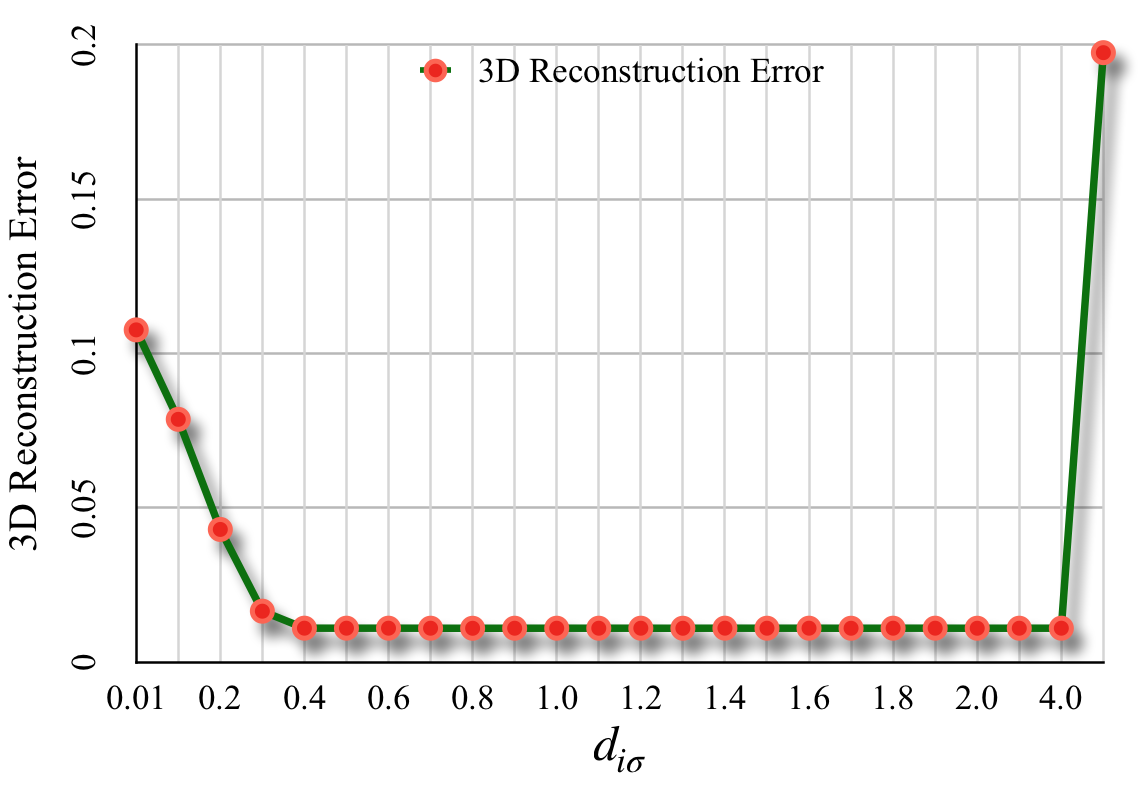}}
\subfigure [\label{fig:sm1}] {\includegraphics[width=0.23\textwidth, height=0.13\textheight]{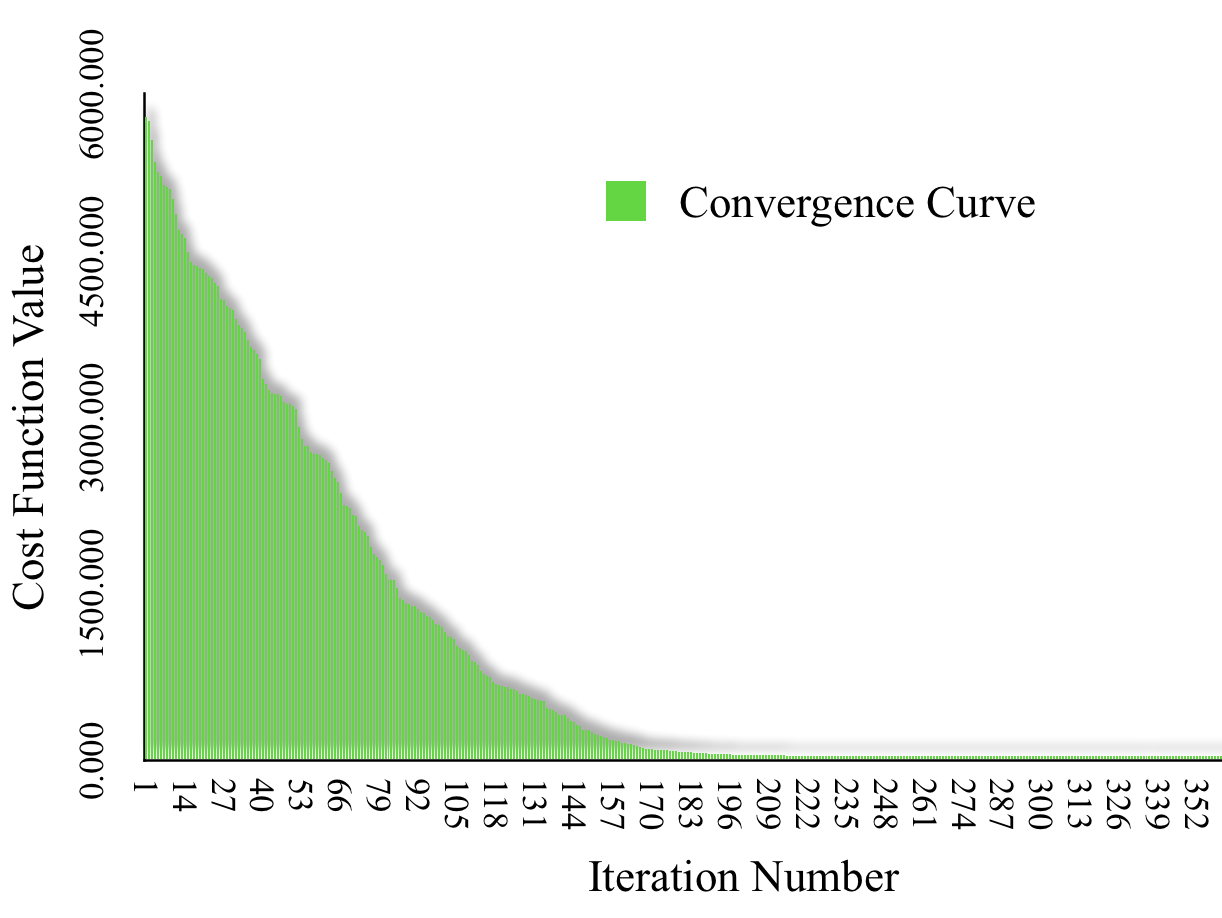}}
\subfigure [\label{fig:sm2}] {\includegraphics[width=0.23\textwidth, height=0.13\textheight]{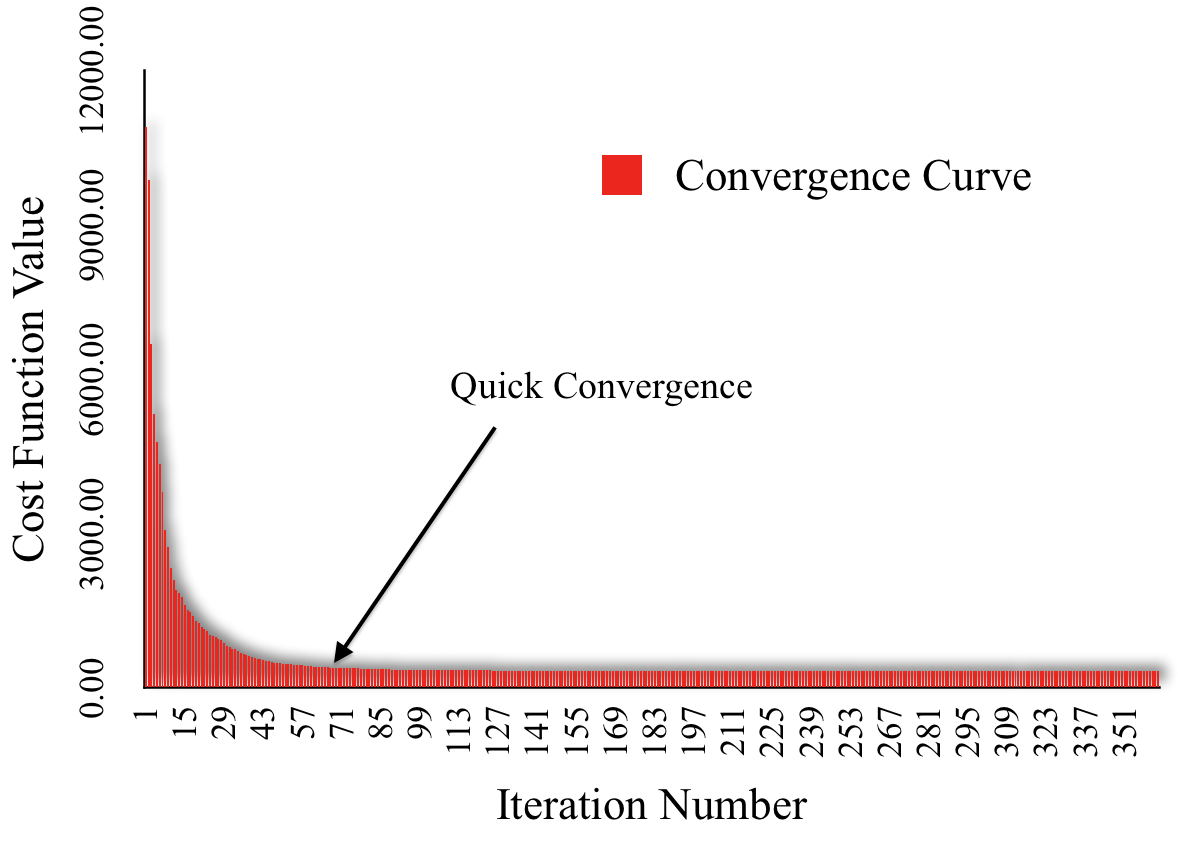}}
\caption{\small{(a) Depth results for the next frame with different levels of Gaussian noise in the reference frame coordinate initialization. (b) Variation in the performance with the change in the $d_{i\sigma}$ values for synthetic example. (c) Convergence curve of the ARAP objective function. (d) Quick convergence with similar accuracy on the same example can be achieved by using restricted isometric constraint.}}
\label{fig:statsexperiment2}
\end{figure*}

\noindent
{\bf{(c) Performance of the algorithm under noisy initialization:}} This experiment is conducted to study the sensitivity of the method to noisy depth initialization. Fig.~\ref{fig:sn1} shows the change in the 3D reconstruction accuracy with the variation in the level of noise from 1\% to 9\%. We introduced the Gaussian noise using randn() MATLAB function and the results are documented for the example shown in Fig.~\ref{fig:syntheticexample} after repeating the experiment for 10 times and taking its average values. We observe that our algorithm can provide arguable results when the noise level gets high.

\begin{figure}
\begin{center}
{\includegraphics[width=0.65\linewidth, height=0.14\textheight]{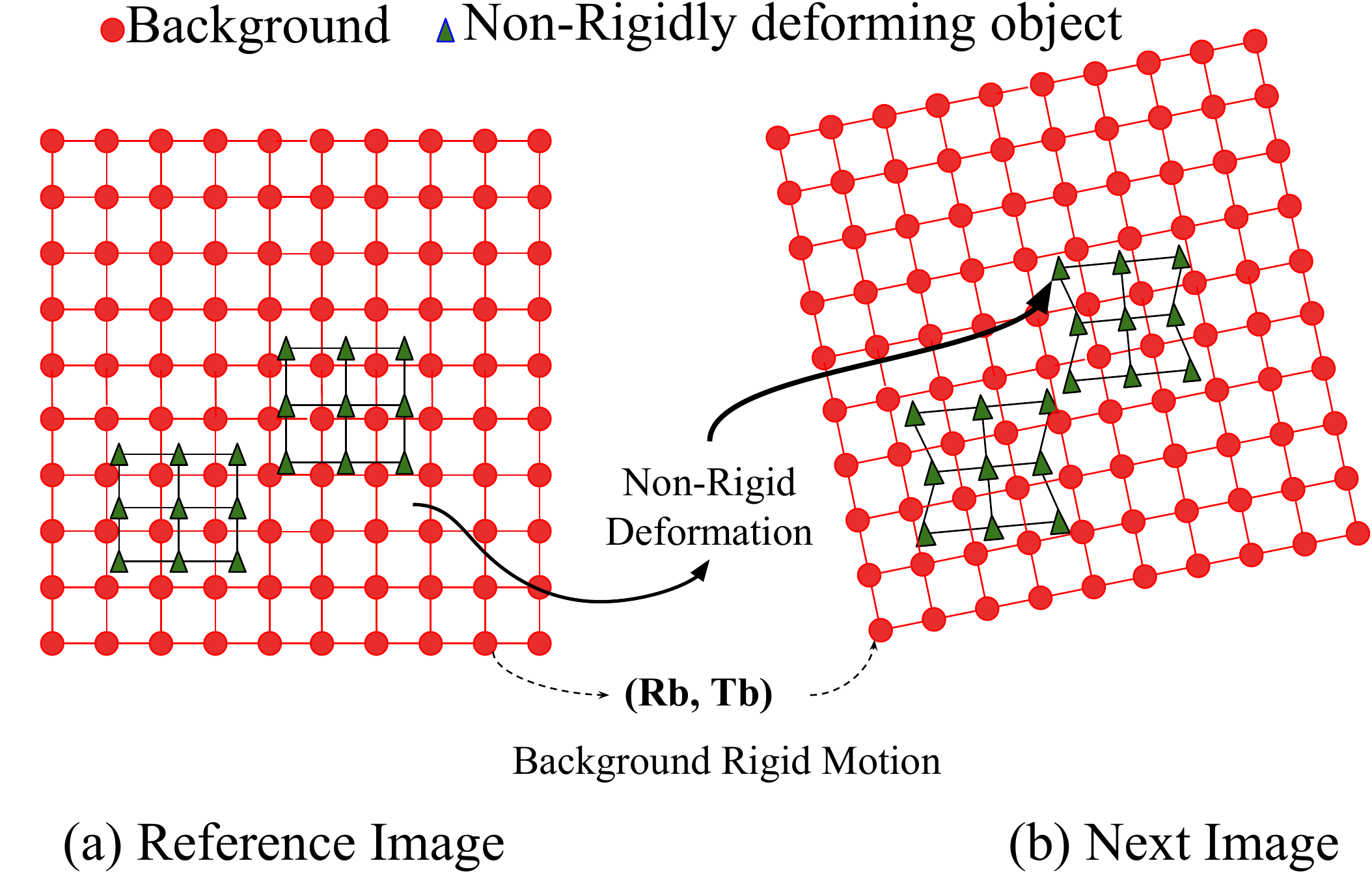}}
\caption{\small{Synthetic example to conduct in-depth behavior analysis of the ARAP. Two objects are deforming independently over a rigid background motion. The objects are at a finite separation from the background. For numerical details on this example, kindly go through the supplementary material.} } \label{fig:syntheticexample}
\end{center}
\end{figure}

\noindent
{\bf{(d) Performance of the algorithm under restricted isometry constraint with $\Phi^\textrm{arap}$ objective function:}} While minimizing the ARAP objective function under the $|\tilde{d}_i-d_i|<d_{i\sigma}$ constraint, we restrict the convergence trust region of the optimization. This constraint makes the algorithm works extremely well ---both in timing and accuracy, if an approximate knowledge about the deformation that the scene may undergo is known a priori. Fig.~\ref{fig:sn2} show the 3D reconstruction accuracy as a function of $d_{i\sigma}$ for the example shown in Fig.~\ref{fig:syntheticexample}. Clearly, if we anticipate the scene transformation a priori, we can get high accuracy in less time. See Fig.~\ref{fig:sm2} which show the quick convergence by using this constraint under a suitable range of $d_{i\sigma}$.

\noindent
{\bf{(e) Nature of convergence of the proposed ARAP optimization:}}\\
\noindent
\textbf{1}) \emph{Without restricted isometry constraint}: As rigid as possible minimization $\Phi^\textrm{arap}$ under the constraint $\tilde{d}_i > 0$ is alone a good enough constraint to provide acceptable results. However, it may take a considerable number of iterations to do so. Fig.~\ref{fig:sm1} shows the convergence curve.

\noindent
\textbf{2}) \emph{With restricted isometry constraint}: Employing an approximate bound on the deformation that the scene may undergo in the next time instance can help fast convergence with similar accuracy. Fig.~\ref{fig:sm2} shows that the same accuracy can be achieved in 60-70 iterations.

Note: The per iteration cost without isometry constraint is : 3.6s, whereas with isometry constrain 1.72s when tested on MPI Sintel dataset image \cite{butler2012naturalistic}.

\section{Limitation and Discussion}
\noindent
Even though our method works well for diverse dynamic scenes, there are still a few challenges associated with the formulation. Firstly, very noisy depth initialization for the reference frame can provide unsettling results. Secondly, our method is challenged by the instant arrival or removal of the dynamic subjects in the scene, and in such cases, it may need reinitialization of the depth. Lastly, well-known limitations such as occlusion and temporal consistency, especially around the regions close to the boundary of the images can also affect the accuracy of our algorithm.

\noindent
{\bf{Discussion:}}
In defense, we would like to state that motion based methods to structure from motion is also prone to noisy data \cite{govindu2006robustness, crandall2011discrete}. Algorithms like motion averaging \cite{venu2015motionavg}, M-estimators and random sampling \cite{torr1997development} are quite often used to rectify the solution.

\noindent
(a) \emph{What do we gain or lose by our approach?}\\
Estimating all kinds of conceivable motion in a complex dynamic scene from images is a challenging task, in that respect, our method provides an alternative way to achieve per pixel depth without estimating any 3D motion. However, in achieving this we are allowing the gauge freedom between the frames (temporal relations in 3D over frames).

\noindent
(b) \emph{Depth results has some blocky effects?}
Few blocky artifacts can be observed in the depth results due to discrete piece-wise planar decomposition of the scene. Although we smooth the solution using TRW-S \cite{kolmogorov2006convergent}, the number of particles for each move is taken as 10 to reduce the convergence time, therefore, some artifacts can be observed.

\section{Conclusion}
The problem of estimating per-pixel depth of a dynamic scene, where the complex motions are prevalent is a challenging task to solve. Quite naturally, previous methods rely on standard relative 3D motion estimation techniques to solve this problem, which in fact is a non-trivial task for a non-rigid scene. In contrast, this paper introduces an alternative way to perceive this problem, which essentially trivializes the 3D motion estimation as a compulsory step. Most of the real-world dynamic scenes if observed closely, it can be inferred that it locally transforms rigidly and globally as rigid as possible. Using such acute observation we propose an algorithm to obtain dense depth estimation under the piece-wise planar approximation of a scene without explicitly solving for 3D motion. Results obtained on the benchmark datasets also validate the competence of our idea. We hope that our idea may open up a new direction of research in the development of modern 3D vision system.


\noindent\textbf{Acknowledgement.}{\footnotesize ~This work is funded in part by the ARC Centre of Excellence for Robotic Vision (CE140100016), ARC Discovery project on 3D computer vision for geo-spatial localisation (DP190102261), ARC DECRA project DE140100180 and Natural Science Foundation of China (61420106007, 61871325)}.


{\small
\bibliographystyle{ieee}
\bibliography{Direct_Dense_Depth}
\nocite{kumar2016multi}
\nocite{kumar2017spatio}
\nocite{kumar2018scalable}
\nocite{kumar2019jumping}
\nocite{kumar2019simple}
}

\end{document}